\documentclass[journal=jacsat,manuscript=article]{achemso}

\usepackage[version=3]{mhchem} 
\usepackage{amsmath, amssymb, amsthm}
\usepackage{graphicx}       
\usepackage{hyperref}       
\usepackage{geometry}       
\usepackage{lipsum} 		 
\usepackage{placeins} 
\usepackage{afterpage} 
\usepackage{amsmath}
\usepackage{booktabs} 
\usepackage{threeparttable} 
\usepackage{placeins}
\usepackage{caption}
\usepackage{booktabs}
\usepackage{threeparttable}
\usepackage{multirow}
\usepackage{enumitem}
\usepackage{xcolor}
\usepackage[normalem]{ulem}

\author{Haocheng Tang}
\affiliation[University of Pittsburgh]
{School of Pharmacy, University of Pittsburgh, Pittsburgh, Pennsylvania 15261, United States}
\email{hat170@pitt.edu}
\author{Jing Long}
\affiliation[Peking University]
{School of Software \& Microelectronics, Peking University, Beijing, 100871, China}
\email{jing.long0926@gmail.com}
\author{Beihong Ji}
\affiliation[University of Pittsburgh]
{School of Pharmacy, University of Pittsburgh, Pittsburgh, Pennsylvania 15261, United States}
\email{bej22@pitt.edu}
\author{Junmei, Wang}
\affiliation[University of Pittsburgh]
{School of Pharmacy, University of Pittsburgh, Pittsburgh, Pennsylvania 15261, United States}
\email{juw79@pitt.edu}

\title[An \textsf{achemso} demo]
  {Auxiliary Discrminator Sequence Generative Adversarial Networks (ADSeqGAN) for Few Sample Molecule Generation}

\abbreviations{ML, GAN, NA, CNS}

\begin{document}
\begin{tocentry}

\centering
\includegraphics[width=9cm,height=3.5cm,keepaspectratio]{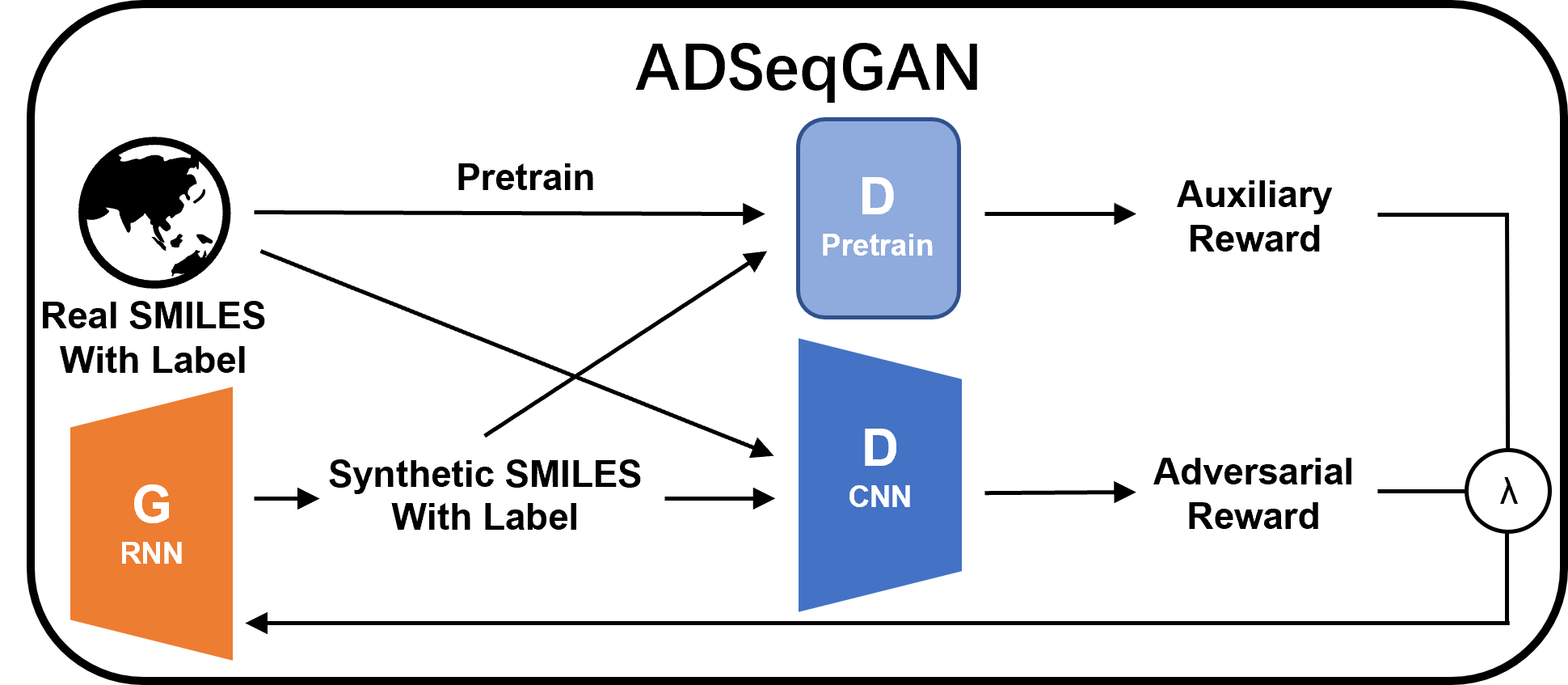}

\end{tocentry}

\begin{abstract}
In this work, we introduce Auxiliary Discriminator Sequence Generative Adversarial Networks (ADSeqGAN), a novel approach for molecular generation in small-sample datasets. Traditional generative models often struggle with limited training data, particularly in drug discovery, where molecular datasets for specific therapeutic targets, such as nucleic acids binders and central nervous system (CNS) drugs, are scarce. ADSeqGAN addresses this challenge by integrating an auxiliary random forest classifier as an additional discriminator into the GAN framework, significantly improves molecular generation quality and class specificity. Our method incorporates pretrained generator and Wasserstein distance to enhance training stability and diversity. We evaluate ADSeqGAN across three representative cases. First, on nucleic acid- and protein-targeting molecules, ADSeqGAN shows superior capability in generating nucleic acid binders compared to baseline models. Second, through oversampling, it markedly improves CNS drug generation, achieving higher yields than traditional de novo models. Third, in cannabinoid receptor type 1 (CB1) ligand design, ADSeqGAN generates novel druglike molecules, with 32.8\% predicted actives surpassing hit rates of CB1-focused and general-purpose libraries when assessed by a target-specific LRIP-SF scoring function. Overall, ADSeqGAN offers a versatile framework for molecular design in data-scarce scenarios, with demonstrated applications in nucleic acid binders, CNS drugs, and CB1 ligands.
\end{abstract}

\section{Introduction}

Non-supervised molecular generation has become a cornerstone of modern computational drug discovery, offering innovative approaches for designing novel compounds with desired properties. Over time, diverse methodologies have emerged, categorized by generative objectives and molecular representations. For instance, models have been developed for molecular property optimization, probabilistic distribution learning, and site-specific design\cite{Du2024}. Molecular representations vary from simplified molecular-input line-entry system (SMILES) strings and molecular graphs to molecular fingerprints and 3D point clouds\cite{Tong2021,Xie2022,Benoit2023}. Meanwhile, neural network architectures like Word2Vec (W2V), sequence-to-sequence (Seq2Seq) models, transformers, graph convolutional networks (GCNs), graph attention networks (GATs), message-passing neural networks (MPNNs), and 3D-point networks have powered these advancements. Generative models such as recurrent neural network (RNN), generative adversarial networks (GANs), variational autoencoders (VAEs), adversarial autoencoders (AAEs), normalizing flows, diffusion models and large language models have further diversified the toolkit for molecular design\cite{Gao2020,Tong2021,Wang2023,Pang2024,Mswahili2024}.

Among all the molecular representation, SMILES notations stand out due to their simplicity, widespread database availability, and extensive tool support. Their sequential representation makes them particularly amenable to natural language processing (NLP) techniques, which further reduce computational and storage costs. This positions SMILES-based approaches as highly advantageous for expanding compound spaces guided by molecular properties.

GANs remain a classic and versatile class of generative models, offering key advantages over VAEs and Diffusion models. By avoiding assumptions of Gaussian priors, GANs are well-suited for datasets with non-Gaussian distributions. Additionally, GANs avoid maximum likelihood estimation(MLE), which, while stabilizing optimization, can constrain generative diversity\cite{Cao2018}. Over the years, many GAN variants have been proposed to address specific challenges in sequence generation,including Sequence GAN (SeqGAN)\cite{Yu2016} and Objective Reinforced GAN (ORGAN)\cite{Guimaraes2017}. SeqGAN leverages policy gradients to optimize sequence outputs, while ORGAN incorporates task-specific rewards to guide generation through reinforcement learning (RL). On the other hand, Auxiliary Classifier GANs (ACGANs)\cite{Odena2016} incorporate class labels into the both generator and discriminator to stabilize training, while pre-trained GAN discriminators\cite{Kumari2021}, ensembles of shallow and deep classifiers, have reduced data requirements in computer vision tasks. However, such mechanisms have not yet been explored for sequence-based molecular generation tasks, leaving a gap in integrating these frameworks for SMILES-based generative models.

For SMILES-based generative models, two primary objectives must be addressed during training: (1) learning the syntactic rules of SMILES notation to ensure valid molecule generation; (2) capturing the structural and functional features of molecules within the dataset. Achieving these goals often requires extensive data and carefully tuned network parameters\cite{Lucic2017,Elton2019}. The scarcity of high-quality datasets for specific drug categories poses a significant challenge. Besides, the length of SMILES strings varies. However, many drug molecules share common features, such as small molecule binders targeting nucleic acids and proteins, which both fall under the broader category of small-molecule therapeutics. Similarly, central nervous system (CNS)-targeted drugs and other non-CNS drugs share overlapping characteristics under the drug discovery framework.

Some methods have been explored to generate new molecules on small datasets based on pre-trained models and molecular scaffolds \cite{Moret2020,Fang2023}, but the results are limited for some data sets with narrow distribution of molecular properties.

In this paper, we introduce a novel approach that integrates a pretrained random forest classifier as a auxiliary descriminator into the SeqGAN framework to improve the quality of SMILES generation. To enhance adversarial training stability, we integrate MLE generator pretraining\cite{Yu2016,Guimaraes2017,Putin2018a} and Wasserstein GANs (WGANs)\cite{Arjovsky2017,Guimaraes2017} into the architecture. To the best of knowledge, this GAN architechture has not been explored so far. Our method demonstrates superior performance in generating nucleic acid-targeting molecules on a nucleic acid and protein mixed dataset, achieving higher verified SMILES rates and yields for nucleic acid binders, compared to models trained exclusively on nucleic acid-targeting datasets.

We also address dataset imbalances by introducing targeted data augmentation strategies. For datasets with extreme biases, such as CNS drug datasets, we employ a novel strategy of over-sampling minority molecules while training on a mixed dataset. This approach significantly increases the generation rate of CNS molecules while maintaining diversity and validity. To further illustrate the universality of ADSeqGAN, we conducted tests using the dataset of cannabinoid receptor type 1 receptor (CB1R) ligand. Overall, ADSeqGAN offers a versatile framework for molecular design in data-scarce scenarios.

Our contributions highlight the synergy between sequence-based GANs and auxiliary classifier techniques in molecular generation and provide a practical framework for addressing dataset imbalances in low-data regimes.

\section{Related Work}

Previous GAN-based models for SMILES sequence generation include SeqGAN and ORGAN. These foundational approaches were later extended with downstream networks such as ORGANIC\cite{Sanchez-Lengeling2017}, RANC\cite{Putin2018a} and ATNC\cite{Putin2018b}, which tailored the generative process to specific application objectives. 

Since the introduction of GANs\cite{Goodfellow2020}, advancements in architectures\cite{Radford2015,Odena2016,Karras2018,Karras2021,Huang2025}, training strategies\cite{Karras2017}, and objective functions\cite{Durugkar2016,Arjovsky2017,Albuquerque2019,Kumari2021,Feng2023} have led to significant progress. Despite their success in image-based tasks, many methods have not yet been applied to GANs for sequence generation. In this work, we integrate these advancements including ACGAN, pre-trained discriminators, and WGAN objective function into a sequence-generation GAN framework, with a particular focus on molecular sequence generation.

Molecules, due to their structured nature and extensive prior knowledge, are particularly well-suited for transfer learning. Descriptor-generation tools like RDKit\cite{rdkit} and OpenBabel\cite{O'Boyle2011} allow for the extraction of rich molecular features, which can be effectively transferred to unseen tasks, datasets, and domains. In our work, we leverage these transferable molecular property representations for unsupervised model training, enabling the generation of high-quality SMILES strings even with limited training data.

GANs can amplify data to address data scarcity issue in molecule predicting task\cite{arigye2020}, outperforming traditional methods like Synthetic Minority Oversampling Technique (SMOTE). At present, data enhancement methods have not been used to generate molecules in GANs. For highly imbalanced datasets, we employ over-sampling to enrich minority classes. 

\section{Methods}


\begin{figure*}[!htb]
    \centering
    \includegraphics[width=0.95\textwidth]{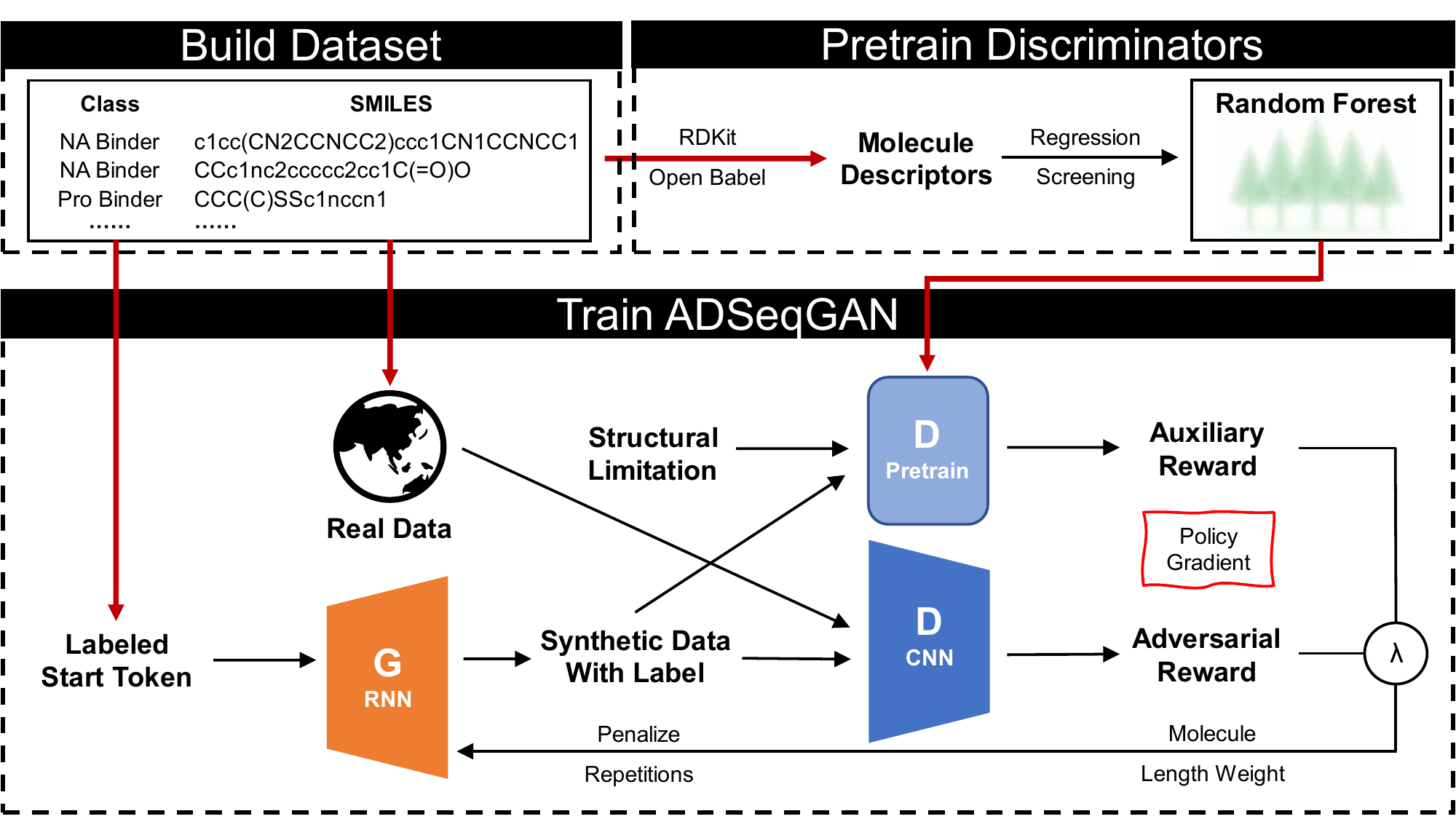}
    \caption{\small Scheme of ADSeqGAN. $Build \ Dataset$: The dataset contains 2 part, including class labels and SMILES strings. It must contains 2 or more different molecule classes. $Pretrain \ Discriminators$: Using RDKit and OpenBabel to calculate molecule descriptors or fingerprints for every datapoints and then choose the descriptors with the strong resolution to build the classifier. To get the pretrained $D$, structural limitation is added. $Train \ ADSeqGAN$: $G$ is fed with labeled start tokens and traind by RL to generate synthetic data of different classes. $D$ with CNN is designed to ditinguish generated data from real samples, while pretrained $D$ is used to classify the samples. Molecules are generated via Monte Carlo sampling. Reward of each token in generated molecules is a linear combination of adversarial and auxiliary reward parameterized by $\lambda$, and is passed back to the policy gradient. We add length weight and repetition penalization to improve quality of new molecules.}
    \label{fig:fig1}
\end{figure*}


Classic GANs consist of two parts: generator $G$, parameterized by $\theta$ to produce sample $Y$ and discriminator $D$, parameterized by $\phi$ to distinguish synthetic data from real ones. The generator and discriminator of original GAN models are trained in alternation, following a minimax game:
\begin{equation}
\begin{aligned}
\min_{G}\max_{D}V(D,G) &= \mathbb{E}_{Y \sim p_{\text{data}}(Y)}[\log D(Y)] \\
&\quad + \mathbb{E}_{Y \sim p_{G_{\theta}}(Y)}[\log (1-D(Y))]
\end{aligned}
\tag{1}
\end{equation}

However, these traditional GANs can not be directly applied to sequence generation tasks, so GAN+RL based SeqGAN and ORGAN were developed. Nonetheless, in our test task, these models struggled to learn enough features with limited real data. To solve these problems, we propose Auxiliary Discrminator Sequence Generative Adversarial Networks (ADSeqGANs). The process of applying ADSeqGANs to molecular generation is shown in Figure 1: Firstly, a hybrid database consisting of a small number of samples of the desirable class and samples of auxiliary classes is constructed. Next the molecule descriptors with strong classification ability are selected by logistic regression method as parameters — or alternatively, molecular fingerprints can be directly employed — for pre-training to get a classifier, which is then added to GAN training as auxiliary discriminators. In parallel, labels of the real data are also entered into the generator to assist in generating samples. Last, we fine-tune the structure and parameters of the network.

Overall, this new network can benefit us in three ways: First, training shallow classifiers on pre-trained features is a common way to adapt deep networks to small data sets while reducing overfitting\cite{Girshick2014,Chen2019,Liang2025}; Second, the discriminator constructed based on prior physical and chemical knowledge may be more in line with human perception\cite{Zhang2018}; Third, adding standardization based on real sample or prior knowledge can reduce mode collapse\cite{Dai2023}.

\subsection{Formulation}
In this work, we introduce a set of molecular features $\mathcal{F}$ and a set molecule classes $\mathcal{C}$, used to train classifiers as aided discriminators ${\{D_n\}}^N_{n=1}$, through corresponding classifier function ${\{C_n\}}^N_{n=1}$. During discriminator training, only adversarial discriminator is updated and the other auxiliary classifiers are frozen. $a_c$ and $b_c$ are structural restrictions based on prior knowledge and certain molecule class $c$. For each auxiliary discriminator, the optimize function is:
\begin{equation}
\begin{aligned}
&\min_{G}V(D_n,G) = \sum_{c \in \mathcal{C}}\mathbb{E}_{Y|c \sim p_{\text{data}}(Y|c)}[\log D_n(Y|c)] \\
&\text{where }D_n = a_cC_n(\mathcal{F})+b_c
\end{aligned}
\tag{2}
\end{equation}

Then the total training is to find the optimal solution of:
\begin{equation}
\begin{aligned}
&\min_{G}\max_{D}\lambda_0 V(D_{\phi},G) + \sum_{n=1}^N\lambda_nV(D_n,G) \\
&\text{where }\sum_{n=0}^N\lambda_n=1
\end{aligned}
\tag{3}
\end{equation}

In this paper, we simply use pretrained random forest as a single auxiliary discriminator. Using more pretrained discriminators is computationally and memory-intensive and does not significantly improve the model's performance.

For discrete dataset, like SMILES strings of molecules, the sampling process is undifferentiable. One successful approach is to train $G_{\theta}$ using a RL model via policy gradient\cite{Williams1992,Yu2016,Guimaraes2017}. Considering a full length sequence $Y_{1:T} = string(t_1t_2...t_T)$, representing a discrete data, $Y_{1:t}$ is an incomplete subsequence belonging to $Y_{1:T}$. One can maximize the long term reward in the policy gradient process to mimic the expectation in $eq(1)$:
\begin{equation}
\begin{aligned}
J(\theta)&=E\left[R_{T} \mid s_{0}, \theta\right] \\
&=\sum_{y_{1} \in Y} G_{\theta}\left(y_{1} \mid s_{0}\right) \cdot Q\left(s_{0}, y_{1}\right)
\end{aligned}
\tag{4}
\end{equation}

where $R_{T}$ is the reward for a complete sequence with the length of $T$ from the discriminator to generator, $s_{0}$ is a fixed initial state at time 0, $y_{1}$ is the next token at time 1, $G_{\theta}\left(y_{1} \mid s_{0}\right)$ is the policy that action $a$ will be taken as state $s_{0}$ to get token $y_{1}$, and $Q(s,a)$ is the action-value function that represents the expected reward at state $s$ of taking action $a$ and following our current policy $G_{\theta}$ to complete the rest of the sequence. According to $eq(2)$ and $eq(3)$, we can decompose $Q\left(s_{0}, y_{1}\right)$ as:
\begin{equation}
\begin{aligned}
Q\left(s|Y_{1: T-1}, a|y_{T}\right)=\lambda_0 R_{\phi,T} + \sum_{n=1}^N\lambda_nR_{n,T} \\
\end{aligned}
\tag{5}
\end{equation}

However, the above equation is only for full sequence $Y_{1: T}$. We also want to get the initial state $s_0$ and $Q$ for partial sequences. $s_0$ cannot directly be used, since the RL processes will consume more computational resources and may reduce the diversity of generated molecules. Instead, $s_x$ is set as the initial state, directly sampling $x$ tokens, as suggested by ORGAN. For the following generation steps, we perform $M$-time Monte Carlo search with the canonical policy $G_{\theta}$ to calculate $Q$ at intermediate time steps. Thus we can evaluate the final reward when the sequence is completed:
\begin{equation}
\begin{aligned}
\operatorname{MC}^{G_{\theta}}\left(Y_{1: t} ; M\right)=\left\{Y_{1: T}^{1}, \ldots, Y_{1: T}^{M}\right\} 
\end{aligned}
\tag{6}
\end{equation}

where $Y_{1: t}^{m}=Y_{1: t}$ and $Y_{t+1: T}^{m}$ is stochastically sampled via the policy $G_{\theta}$. Now $Q(s, a)$ becomes $Q(t)$:
\[
Q(t) = 
\begin{cases} 
\frac{1}{M} \sum_{m=1}^{M} (\lambda_0 R^m_{\phi,T}+\sum_{n=1}^N\lambda_nR^m_{n,T})  \\
\text{with } Y_{1:T}^{m} \in \mathrm{MC}^{G_{\theta}}\left(Y_{1:t}; M\right), \\
\quad\quad\quad\quad\quad\quad\quad\quad\quad\quad\quad\quad \text{if } t < T. \\ 
\lambda_0 R_{\phi,T}+\sum_{n=1}^N\lambda_nR_{n,T}, \quad \text{ if } t = T.
\end{cases}
\tag{7}
\] 

Finally, according to SeqGAN, we can get an unbiased estimation of the gradient $J(\theta)$:

\begin{equation}
\begin{aligned}
\nabla_{\theta} J(\theta) \simeq \frac{1}{T} & \sum_{t=1, \ldots, T} \mathbb{E}_{y_{t} \sim G_{\theta}\left(y_{t} \mid Y_{1: t-1}\right)} \\ &[\left.\nabla_{\theta} \log G_{\theta}\left(y_{t} \mid Y_{1: t-1}\right) \cdot Q(t)\right] 
\end{aligned}
\tag{8}
\end{equation}
\subsection{Implementation Details}
$G_{\theta}$ is a RNN with long short-term memory (LSTM) cells for sequence generation, while $D_{\phi}$ is a convolutional neural network (CNN) for text classification tasks\cite{Kim2014}. Unlike SeqGAN and ORGAN, ADSeqGAN uses labeled start token as input, and different classes share the same RNN.

To avoid problems of GAN convergence like "perfect discriminator" and improve the stability of learning, we introduce the Wasserstein-1 distance, also known as earth mover's distance\cite{Gulrajani2017}, to $D_{\psi}$. In essence, Wasserstein distance turns the discriminator's classification task into a regression task, with the goal of reducing the distance between the real data distribution and the model distribution, which is more in line with the goals of RL.

To guide the generator toward producing structurally reasonable and class-specific molecules, we incorporated a pretrained classifier-based discriminator $D_n$ as an auxiliary scoring module. First, we computed 192 molecular descriptors for all molecules using RDKit and OpenBabel, covering physicochemical, topological, and electronic properties. We then selected a subset of features showing high discriminative power, based on logistic regression and their AUC performance in 5-fold cross-validation, to train a random forest classifier using 100 trees. The resulting model was saved and deployed during generation scoring. Specifically, the probability of being correctly classified as ground truth label is given as an initial reward value. 

In addition to classifier prediction, we implemented a rule-based filtering system to enforce physically motivated constraints. These include penalties for excessive nitrogen/oxygen/sulfur content, macrocyclic ring systems, and anti-aromatic rings, as well as rejection of molecules with high halogen ratios or extremely carbon-dominant structures. Conversely, positive scoring is applied to fused aromatic systems with suitable heteroatoms. The final discriminator score is computed as the product of the classifier probability and the physics-based structural score. Molecules with invalid SMILES, extreme topology, or poor drug-likeness are thereby suppressed, ensuring that generated outputs are both class-relevant and chemically meaningful.

Other additional mechanisms to prevent mode collapse and over-fitting includes: (1) Using $0-1$ standardization in rollout policy when computing rewards. (2) Penalizing repeated sequences by dividing the reward of a non-unique sample by the number of copies. (3) Applying length weight to make sure the length of molecules generated correspond the real length distribution of samples.

\subsection{Dataset Preparation}

We constructed two datasets for evaluating our model performance: (1) small molecules targeting nucleic acids or proteins, and (2) central nervous system (CNS) and non-CNS drugs.

\textbf{Nucleic Acid and Protein Binder Dataset.}  
This dataset was curated from ChEMBL345 \cite{Gaulton2012,Zdrazil2024} and all publicly available small-molecule structures on DrugBank \cite{Knox2023}. Molecules were selected based on annotations indicating their interaction with either nucleic acids or proteins.

\textbf{CNS vs. Non-CNS Drug Dataset.}  
CNS and non-CNS drugs were collected from DrugBank, including all approved and investigational small-molecule drugs with known CNS indications up to December 2024.

For both datasets, we applied the following preprocessing steps using RDKit:
\vspace{1pt}
{
\begin{itemize}[itemsep=1pt, topsep=0pt, parsep=0pt, partopsep=0pt]
    \item Removal of 3D structural information and inorganic compounds.
    \item Exclusion of molecules with excessive length or containing elements beyond C, H, N, O, halogens, B, S, and P.
    \item Removal of salts and inorganic ionic components.
    \item Canonicalization of SMILES strings for uniformity.
\end{itemize}
}

After preprocessing, the nucleic acid binder dataset contains 4,894 molecules, while the CNS drug dataset contains 548 CNS-targeting molecules approved and entering clinical trial before December 2024. The distributions and molecular properties of the datasets are summarized in Figure~S5.

\subsection{Docking}

For nucleic acid binders, we performed local flexible docking of generated molecules to target nucleic acids using NLDock v1.0 \cite{Feng2021}. NLDock has demonstrated superior local flexible docking accuracy over major docking tools to analyze nucleic acid and ligand interactions. First, \texttt{BDS} function in NLDock was used to define spherical binding‐site points from the receptor PDB. For each ligand, up to 100 conformations were generated using RDKit's ETKDG method and converted via OpenBabel to MOL2 format. NLDock then docked these conformers to the target using the generated sphere points, retaining a maximum of 50 poses per ligand. Conformations were ranked by NLDock’s native scoring using \texttt{SortConfs}, and RMSD to a native ligand was calculated with \texttt{obrms}. 

For CNS drugs, Molecular Operation Environment (MOE) 2019.0102 was used to perform molecular docking \cite{Vilar2008}. MOE is a widely used commercially available docking software. Triangle matcher method based on London dG score was used to place ligands at the pocket site to generate 30 poses. For refinement, we used rigid receptor and GBVI|WSA dG score to get top 5 binding poses of each molecule.    

\section{Results}

\renewcommand{\arraystretch}{1.5} 
\begin{table*}[h]
    \centering
    \scriptsize
    \begin{threeparttable}
	    \caption{Evaluation of metrics on several generative algorithms to generate nucleic acid binders}
	    \label{tab:first}
	    \begin{tabular}{l c cc c c c c c c}
	        \toprule 
	        \multirow{2}{*}{Algorithm} & \multirow{2}{*}{Dataset} & \multicolumn{2}{c}{Novelty/\%} & \multirow{2}{*}{$\text{SA}^{\Uparrow}$} & \multirow{2}{*}{$\text{QED}^{\Uparrow}$} & \multirow{2}{*}{$\text{FCD}^{\Downarrow}$} & \multirow{2}{*}{NA ratio/\%} & \multirow{2}{*}{NA yield/\%} & \multirow{2}{*}{Size} \\  
	        \cmidrule(lr){3-4}
	                 &         &\footnotesize Validity/\% &\footnotesize Unique/\% &  &  &  &  &  &  \\  
	        \midrule 
	        MLE RNN          & NA       & 17.7 & 99.3 & 0.53 & 0.58 & 0.90 & 17.6 & 3.1 &6400\\ 
	                         & NA+Pro   & 19.3 & 97.1 & 0.44 & 0.53 & 0.81 & 3.8  & 0.7 &6400\\ 
	        Native RL        & NA       & 68.2 & 99.2 & 0.30 & 0.51 & 0.74 & 29.8 & 20.1&6400\\ 
	                         & NA+Pro   & 51.7 & 99.4 & 0.02 & 0.35 & 0.72 & 1.2  & 0.6 &6400\\ 
	        SeqGan           & NA       & 0    & --   & --   & --   & --   & --   & --  &6400\\ 
	        ORGAN            & NA       & 1.3  & 98.8 & 0.01 & 0.44 & 0.64 & --   & --  &6400\\ 
	                         & NA+Pro   & 0    & --   & --   & --   & --   & --   & --  &6400\\   
	        MolGPT           & NA       & 0.4  & 22.0 & 0.00 & 0.014& 0.49 & --   & --  &10000\\ \hline	        
	        $\text{CLM}^*$   & NA       &\multicolumn{2}{c}{8.8} & 0.21 & 0.61 & 0.81 & 31.7 & 2.9 &10000\\  
	        $\text{MolGen}^*$& NA       &\multicolumn{2}{c}{1.6} & 0.17 & 0.21 & 0.86 & 38.8 & 0.6 &924800\\ \hline
	        ADSeqGAN         & NA+Pro   & 94.5 & 71.6 & 0.78 & 0.40 & 0.66 & 82.7 & 56.0&6400\\
	        \bottomrule
	    \end{tabular}
	    \begin{tablenotes}	
			\item[*] Using pretrained models.
			\item[] \hspace{-1em} $\Uparrow$ means the larger the better, while $\Downarrow$ means the smaller the better.
	    \end{tablenotes}
	\end{threeparttable}
\end{table*}

\renewcommand{\arraystretch}{1} 

Here we conduct experiments to test ADSeqGAN in two representative scenarios: a moderately biased dataset consists of 4894 nucleic acid binders (NA) and 8191 protein targeting molecules (Pro) with high bioactivity, and an extremely biased dataset consisting of 548 CNS drugs and 7728 other small molecule drugs. Our objective is to prove that our model can generate functional drug molecules with only a small number of target samples while promoting synthesizability and drug-likeness, diversity and similar molecule length distribution. To estimate the quality of generated samples, score functions designed for synthesizability (SA)\cite{Ertl2009}, quantitative estimate of drug-likeness (QED)\cite{Lipinski2001,Bickerton2012,Ivanović2020} and Frechet ChemNet Distance (FCD)\cite{Preuer2018} are used, and we use fingerprint ECFP4\cite{Rogers2010} to compute Tanimoto similarity\cite{Nikolova2003} for diversity evaluation.

During training, every generator model was pretrained for 250 epochs based on MLE, and the discrminator was pre-trained for 10 steps. Unless otherwise stated, all GAN models use Wasserstein distance. Unless specified, $\lambda$ is set to 0.2. MLE-based RNN, SeqGAN, RNN-pretrained RL, chemical language model (CLM)\cite{Moret2020}, MolGPT\cite{Bagal2022}, MolGen\cite{Fang2023} are selected for contrast experiments. All GAN-related models and RL-based models are set to be pretrained for 250 epochs of MLE-based $G_{\theta}$, and then 50 epochs for downstream training. Meanwhile, in pretraining steps of GANs, $D_{\psi}$ is set to be pretrained for 10 steps. For other models, the training parameters are based on original papers. MolGen is used to generate new molecules based on the backbone of each single molecule, so 200 new molecules were generated per input. Notably, for all drug molecules in GAN and RL relative models, grammar restrictions are loose and the later generated samples are dealt with openbabel's gen2d function to reconstruct strictly valid molecules, since lightly invalid SMILES are beneficial rather than detrimental to chemical language models\cite{Skinnider2024}. For all generative models, the generation length was constrained to the range of 10 to 80 tokens to ensure chemical validity and relevance.

\subsection{Nucleic Acid Binder Generation}

Nucleic acid-targeting drugs remain relatively scarce compared to their protein-targeting counterparts, despite the increasing recognition of nucleic acids as crucial therapeutic targets. In contrast to nucleic acid drugs, nucleic acid binders, though still not so many, are much more diverse and have been widely studied for their interactions with DNA and RNA. These binders play essential roles in regulating gene function and dying nucleic acids. Meanwhile, small molecules and protein-based drugs with high bioactivity have been extensively developed for protein targets, demonstrating well-characterized pharmacological properties.

Given the limited number of nucleic acid binders and the established chemical diversity of small molecules targeting proteins, it is of significant interest to explore a molecular generation model that integrates the chemical characteristics of both. The primary goal is to construct a generative model which is capable of learning the distinct yet complementary features of nucleic acid binders and bioactive small molecules that target proteins. By capturing the underlying chemical space shared between these two classes, such a model could facilitate the discovery of novel nucleic acid-targeting molecules with enhanced specificity and efficacy.

\subsubsection*{Build Pretrained Discriminators}

To distinguish nucleic acid binders from protein-targeting small molecules, we computed molecular descriptors for a dataset of compounds and evaluated their discriminatory power using logistic regression. Following an automatical regression-based analysis, we manually selected 18 descriptors with distinct physicochemical properties that showed relevance in distinguishing these two classes from the top 50. A random forest classifier trained on these selected descriptors achieved an AUC of 0.91 (Figure S1), indicating strong classification performance. 

Furthermore, we have integrated MACCS fingerprint\cite{Campbell2025} based classifer into the pipeline, providing user-friendly choices and it shows slightly better performance than RDKit descriptors in Figure S5. MACCS is also more convenient, requiring no manual feature selection. We also tested deep learning–based descriptors (CDDD) \cite{Winter2018}, which can be directly extended to other chemical spaces, as shown in Figure S5. While CDDD provides the advantage of avoiding manual feature selection and offers broader transferability, it also introduces practical limitations, including compatibility issues, longer loading times for pretrained parameters, and increased computational cost. Therefore, for the current work, we retained RDKit/OpenBabel descriptors in the workflow in ADSeqGAN.

Based on the classifier, we further developed a scoring function, integrating multiple molecular properties, including atomic composition, ring systems, and physicochemical descriptors, to assess molecular fitness. The final score is obtained by combining classifier prediction with the structural penalties, ensuring that molecules adhere to predefined chemical criteria.

\begin{figure*}[!htb]
    \centering
    \includegraphics[width=0.95\textwidth]{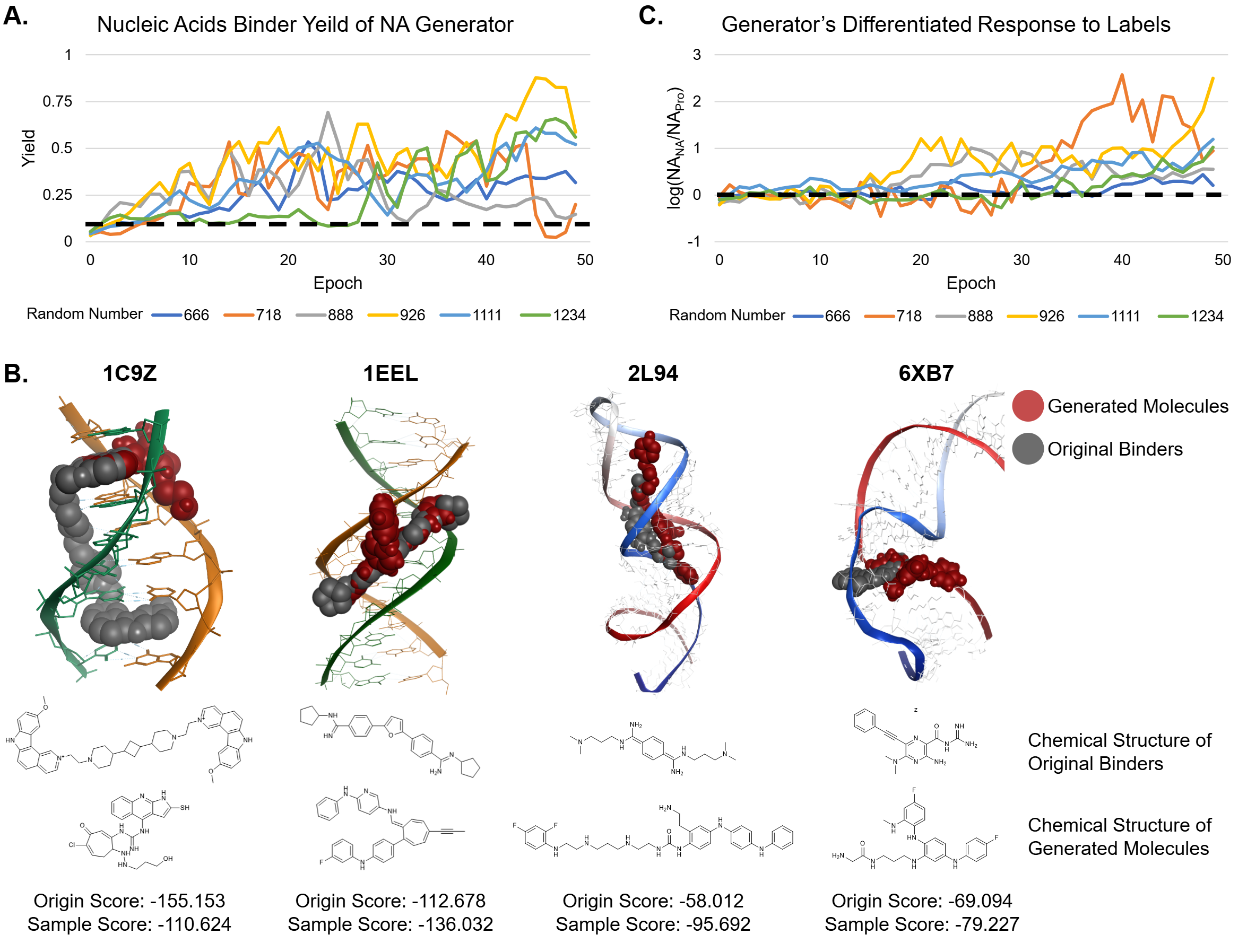}
    \caption{\small ADSeqGAN training results on the nucleic acid and protein dataset. $A.$ The yield of nucleic acid binders using "NA" as the input label changes with the increase of epoch. Yield is calculated by $unique\_ratio\times verified\_ratio \times NA\_ratio$. $B.$ NLDock docking results of generated molecules with various nucleic acid targets. 1C9Z and 1EEL are DNA targets, while 2L94 and 6XB7 are RNA targets. The grey balls are original binders with native conformations, and the red ones are generated samples. $C.$ Plot of label responsiveness with the increase of epoch. The responsiveness metric is calculated by taking the $Log10$ of the NA binder yield ratio, with the numerator as the yield of NA binder after entering "NA" and the denominator as the yield after entering "Pro". }
    \label{fig:fig2}
\end{figure*}

\subsubsection*{ADSeqGAN Performs Best on The Nucleic Acid Binder Generation Task}

In this study, we compare our ADSeqGAN model with representative classic SMILES-based generative models (Table 1). The reference models, including MLE RNN, Native RL, SeqGAN, ORGAN, and MolGPT, were all trained from scratch on the dataset, while CLM and MolGen utilized pre-trained models and parameters to do downstream fine-tuning without incorporating predefined conditions or molecular scaffolds. Under the given experimental conditions, ADSeqGAN outperforms the other non-pretrained models in terms of sample generation success rate, synthetic feasibility, and yield of nucleic acid-targeting molecules. These non-pretrained models struggled to produce a substantial number of small molecules targeting nucleic acids. On the other hand, while testing nucleic acid binder generation task on pretrained models, the generated samples did not meet the corresponding expectations either, especially for NA yield. As training progresses, CLM improves in generating the correct types of molecules, but its validity ratio declines. In contrast, MolGEN exhibits strong responsiveness to certain molecules, producing a diverse set of structures, while for others, it fails to differentiate and repeatedly generates identical structures. 

We then randomly test different random number seeds to observe the training process (Figure 2A). Although the performance was different, the trend showed that the NA Generator gradually learned the characteristics of nucleic acid molecules, and the optimal yield could all be greater than 50\%, far higher than the representative baseline models.  

Additionally, ADSeqGAN achieved impressive FCD scores, indicating that our model learned richer molecular structures, especially comparing with pretrained models. Specifically, the generated nucleic acid binders captured some features of protein-targeting small molecule. As shown in principal component analysis (PCA) in Figure S2, the distributions of the characteristic molecular fingerprint fragments of generated NA samples scattered around the Pro molecular database, rather than just the NA binder dataset. Besides, the introduction of repetition penalties also contribute to more diverse molecular sampling. 

The QED scores were lower compared to pre-trained models, MLE RNN, and some RL-based models. It is worth noting that both MLE RNN and Native RL models, trained on a mixed dataset, exhibited lower QED scores compared to those trained exclusively on nucleic acid data. This suggests that the mixed dataset increased the difficulty of learning QED. Moreover, QED correlates with molecular mass. For instance, the average length of molecules generated by Native RL, which was trained on nucleic acid data alone, was 30.7, whereas ADSeqGAN generated molecules with an average length of 52.8. The higher QED values in CLM and MolGEN are attributed to their pre-training on highly bioactive molecules, which also contributed to their larger QED scores.

\begin{figure*}[!htb]
    \centering
    \includegraphics[width=0.95\textwidth]{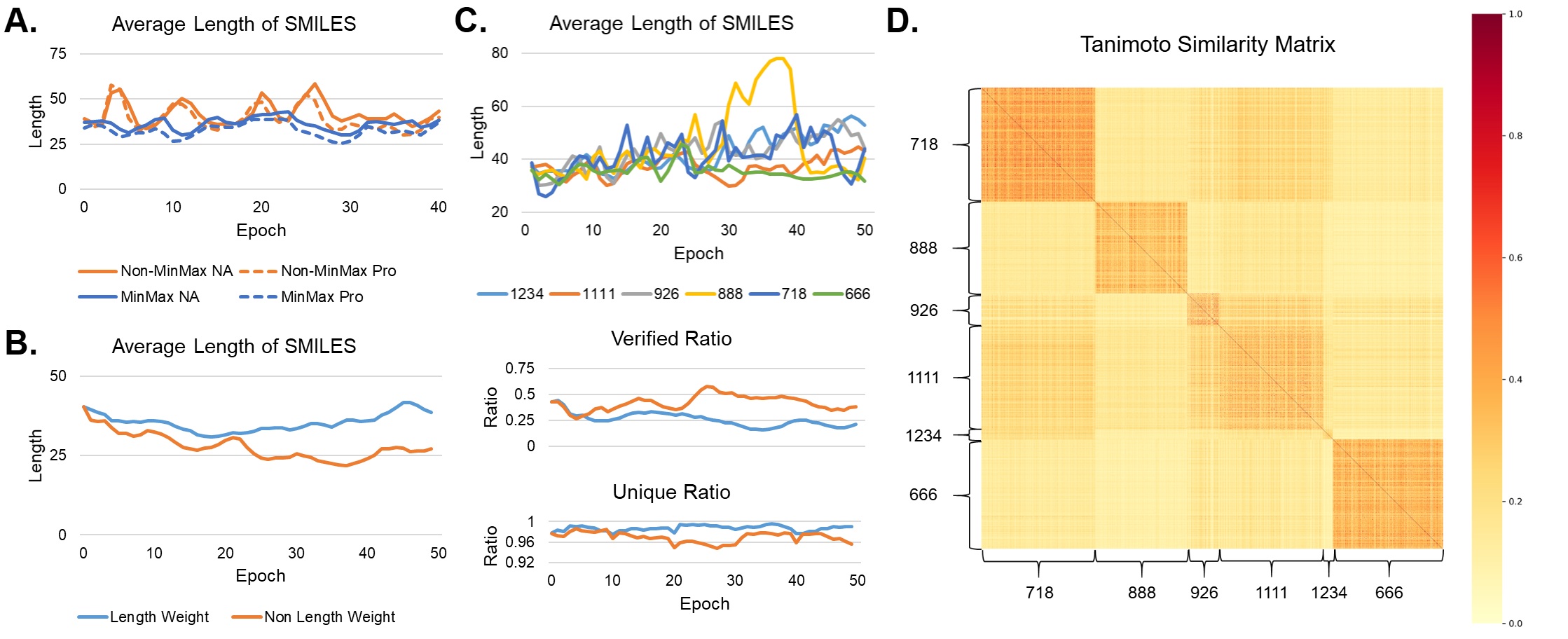}
    \caption{\small Ablation experiments. $A.$ Effect of MinMax regularization on molecular length during training. Orange indicates without, blue indicate with MinMax regulation. The solid line indicates NA and the dotted line indicates Pro. $B.$ The effect of length weight on the length, verified ratio and unique ratio of the molecules generated during training running under the SeqGAN framework. $C.$ The effect of random number on molecular length during training. $D.$ Tanimoto similarity matrix of samples generated by different random numbers at the 40th epoch.}
    \label{fig:fig3}
\end{figure*}

To further validate the effectiveness of the generated molecules, we selected four DNA targets (1C9Z\cite{Shui2000}, 1EEL\cite{Mazur2000}, 3U05\cite{Wei2013}, 6AST\cite{Harika2017}) and four RNA targets (1UUD\cite{Davis2004}, 2L94\cite{Marcheschi2011}, 2LWK\cite{Lee2014}, 6XB7\cite{Davila-Calderon2020}) for docking experiments using NLDock\cite{Feng2021}, which is specially developed for simulate the interaction between nucleic acids and ligands. The docking process was configured in the local and flexible mode. We used virtual screening datasets, with each consisting of 6,400 molecules generated from a well-trained generator and then chose 2 hits for each target as cases (Figure 2B, S6A and Table S1, S2). The results showed that for seven out of the eight targets, many of our generated molecules exhibited stronger binding affinities than the original binders in their native conformations. In addition, compared with the original molecule, the structures are very diverse. For 1C9Z, its unique pocket shape made it hard for generated molecule to perform better than ground truth. However, using only the native ligand’s SMILES (without experimental binding conformation), we successfully generated molecules—despite the difficulty of matching the pocket shape—that even showed higher affinity than the native ligand. AlphaFold3 was further used to confirm binding poses and observe local interactions. The coincidence of conformations predicted by AlphaFold3 and NLDock further enhances the reliability of the prediction results.

To further understand structural insight of generated molecules, we expanded our analysis to include scaffold distributions and functional group diversity as shown in Figure S7. Nucleic acid binders tend to contain a higher proportion of nitrogen and oxygen atoms and more fused aromatic ring systems, while protein binders more frequently feature aliphatic ring systems. Importantly, the molecules generated by ADSeqGAN captured these preferences, particularly the enrichment of aromatic rings and nitrogen atoms in nucleic acid binders.

It is worth noting that, compared to real molecules, the model struggles to generate highly symmetrical structures, such as those found in 1C9Z, 1EEL, and 2L94. Additionally, certain targets, such as 1PBR \cite{Fourmy1996} and 1QD3 \cite{Faber2000}, have small-molecule ligands that entirely lack aromatic rings. However, in our experiments, we did not observe any generated samples without aromatic rings or double bonds. This limitation may stem from the scarcity of such molecules in the training dataset, making it difficult for the model to learn their characteristics. The proportion of fused aromatic rings tends to decrease because the SMILES notation for such structures is more complex and difficult for GAN and RL to learn. Actually, we find molecules with fused aromatic rings targeting 1C9Z most in the first 25 epochs. Only certain random seeds (e.g., 666 and 888 after 40 reinforcement learning steps) yielded models capable of producing fused-ring structures. This reflects an inherent limitation of combining GANs with reinforcement learning, which may converge to local optima and struggle to capture highly complex structural features.

\subsubsection*{ADSeqGAN Differentially Generates Nucleic Acid and Protein Binders}

ADSeqGAN generates distinct molecular outputs depending on the input label. By calculating the $Log10$ ratio of the proportion of nucleic acid binders among valid molecules when given the NA label to the proportion of nucleic acid binders when given the Pro label at each epoch, we observe that the generator exhibits strong label responsiveness. Specifically, when the NA label is provided, the model preferentially generates nucleic acid binders, whereas inputting the Pro label results in a higher proportion of protein binders. This trend becomes increasingly pronounced as training progresses (Figure 2C).

Furthermore, analyzing the SMILES sequence lengths at each epoch reveals a notable difference in molecular size. As shown in Figure S3, molecules generated with the Pro label tend to be shorter than those generated with the NA label. This pattern is consistent with the length distribution observed in the real dataset, which is 40.4 for Pro and 46.1 for NA, further validating the model's ability to learn and replicate intrinsic structural characteristics of nucleic acid and protein binders.

\subsubsection*{Standardization Reduce Mode Collapse and Over-fitting}

GANs are prone to mode collapse and overfitting during training, often resulting in repetitive sequences such as "c1ccccc1" and "NCCCNCCCN." Unlike traditional SeqGAN and ORGAN, we applied a min-max transformation to map the rewards from both $D_{\psi}$ and $D_n$ to the range [0,1], preventing excessive rewards for certain sequences or bias toward a single discriminator (Figure 3A). This adjustment stabilized the training process, making the length fluctuate less.

\subsubsection*{Length Weight and Penalizing Repetition Give Higher Generation Quality}

Although the authors of ORGAN claimed that GAN-generated molecules exhibit similar lengths to those in real molecular datasets, our experiments revealed that without length penalties, the generated molecules tend to be shorter than those in our dataset. We replicated their findings on the QM9\_5K dataset, where the average SMILES length is only 15.4. However, in our NA+Pro dataset, the average length is 42.8, with more complex structural expressions. As molecular length increases, the success rate of generating valid molecules decreases. Additionally, since the model learns molecular syntax by rewarding only valid molecules, this further biases the generation toward shorter sequences. A similar trend is observed in Native RL trained on NA data, where the final generated molecules have an average length of only 32.7—significantly shorter than the NA dataset’s original average of 46.1.

To address this, we applied both length weighting and repetition penalization during training. As shown in Figure 3B, this resulted in more stable sequence lengths, with generated molecules maintaining an average length around 40. Notably, longer generated molecules exhibited a higher proportion of unique samples and a lower proportion of verified SMILES, suggesting a positive correlation between sequence length and molecular diversity and a negative correlation between sequence length and validation. Therefore, we propose adjusting training parameters dynamically to further enhance generation quality: gradually increasing the weight of length constraint while reducing repetition penalization as the model generates longer SMILES sequences. 

\subsubsection*{Random Number Leads to Diverse Molecules}

RL is highly sensitive to random seeds, with different seeds potentially leading to vastly different results\cite{Henderson2017, Islam2017}. By comparing samples generated from different random seeds (Figure 3C,D), we observe that while all molecules are classified as nucleic acid binders, their specific structures vary significantly. To achieve greater molecular diversity, we strongly recommend conducting experiments with multiple random seeds to obtain a broader range of generated molecules. As shown in Figure S7, seed 1111 produced scaffolds with few oxygen atoms and occasional macrocycles and seed 888 generated more linear structures, while seed 1234 yielded highly branched scaffolds. These findings are consistent with the structural diversity patterns shown in Figure 3D.

\subsection{CNS Drug Generation}

\begin{figure}[!htb] 
    \centering
    \includegraphics[width=0.8\linewidth]{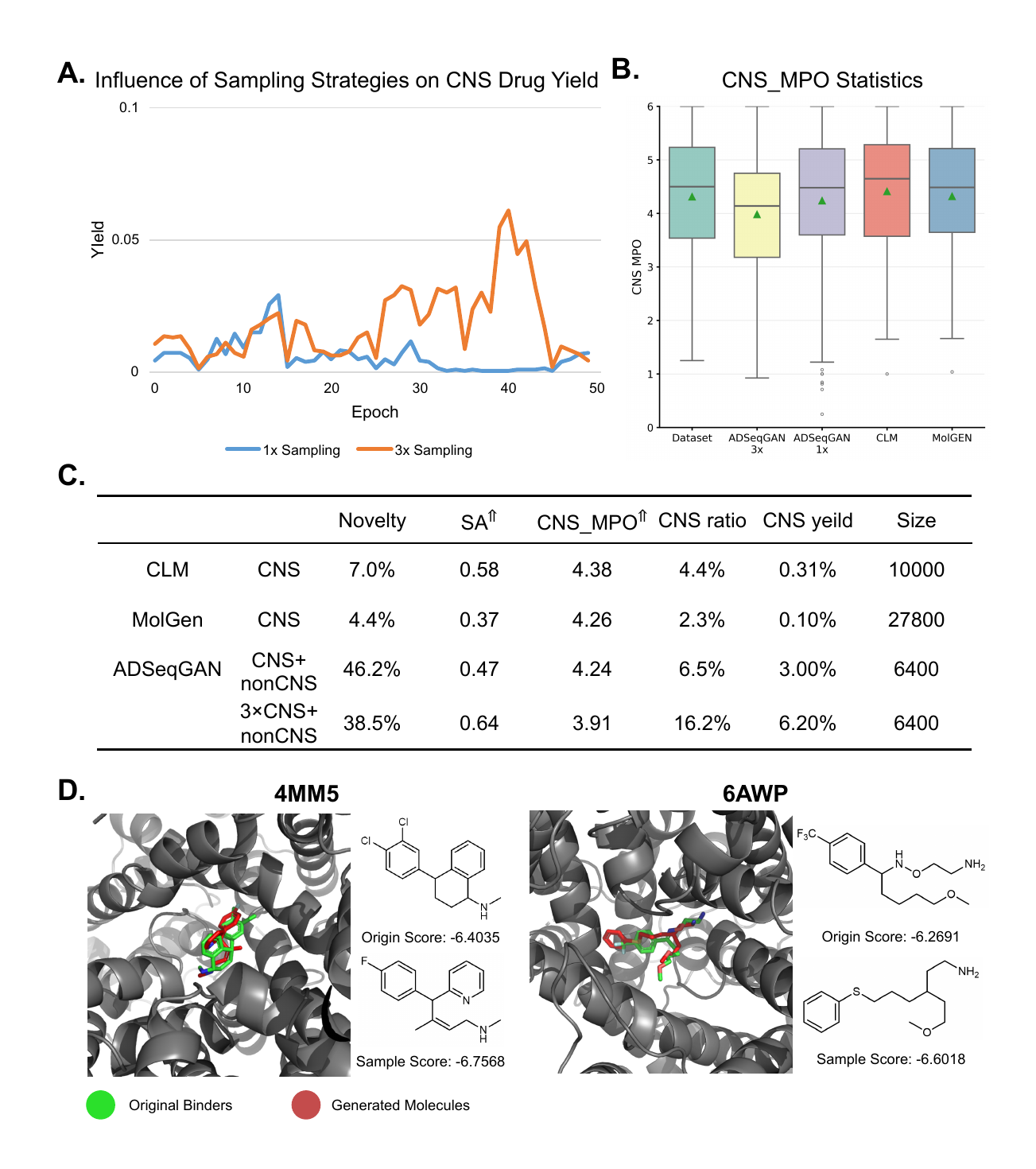}
    \caption{\small ADSeqGAN training results on the CNS and non-CNS drug dataset. $A.$ Influence of sampling strategies on CNS drug yield. $B.$ CNS\_MPO statistic results of original dataset and four few-shot moleculr generation models. $C.$ Evaluation of metrics on few-shot generative algorithms to generate CNS drugs. $\Uparrow$ means the larger the better. $D.$ MOE docking result of generated molecules with various CNS targets. Green sticks show original molecules, while red sticks show generated samples. 4MM5 is LeuBAT (delta13 mutant) in complex with sertraline. 6AWP is ts3 human serotonin transporter complexed with fluvoxamine. }
    \label{fig:fig4}
\end{figure}

Despite the pressing need, the number of approved CNS drugs remains limited. One major obstacle in CNS drug development is the blood-brain barrier (BBB), a selective membrane that restricts the entry of many compounds into the brain, complicating the delivery of therapeutic agents. Recent advancements in artificial intelligence and machine learning offer promising avenues to overcome these hurdles by enabling the design of novel compounds with optimized properties for CNS activity, but the few samples are still a problem\cite{Morofuji2020}. CNSMolGen\cite{Gou2024} is developed for CNS drug generation using a pretrained dataset based on either the central nervous system multiparameter optimization (CNS\_MPO) score \cite{Wager2010, Wager2016} or high bioactivatity, followed by a fine-tuning module focusing on specific drug class. Our model goal is to build on the foundation of drugs that have entered clinical trials and have already been approved. We demonstrate the potential of our small sample based generative model to expand the library of candidate molecules for CNS drugs. To estimate the quality of generated samples, score functions designed for synthesizability (SA),and CNS\_MPO are used. It should be noted that the calculation of CNS\_MPO depends on basic pKa. To facilitate large-scale computing, we used MolGpKa\cite{Pan2021} based on graph neural networks for batch pKa prediction.

To evaluate performance under extremely few-shot conditions for CNS drug generation, we conducted comparative experiments using ADSeqGAN, CLM, and MolGen, all of which support molecular generation in low-data regimes. The training configurations for all models were kept identical to those used in the nucleic acid binder generation task. For MolGen, 50 new molecules were generated per input molecule in the CNS-targeting task.

\subsubsection*{Pretrain Auxiliary Descriminators}

Overall, we adopted the same pipeline as previously used for constructing the nucleic acid-based auxiliary discriminator. Specifically, we first selected 12 molecular descriptors with strong physicochemical discriminability via logistic regression, and then trained a random forest classifier. The resulting classifier achieved an AUC of 0.81 (Figure S4), which is notably lower than the 0.91 obtained for the nucleic acid binder task. This performance gap can be attributed to two factors: (1) the physicochemical properties of CNS and non-CNS drugs are more similar than those between nucleic acid binders and approved drugs; and (2) the number of CNS drug samples is relatively limited. Therefore, directly using this classifier as a discriminator poses certain limitations. To enhance the model's ability to capture CNS-specific characteristics, we adjusted the reward threshold such that a molecule is assigned a reward value of 1 if the classifier predicts a CNS probability greater than 40\%.

\subsubsection*{Performance of ADSeqGAN to Generate CNS Drugs}

Due to the limited availability of CNS drugs in the dataset (only 548 samples), their contribution to the parameter updates during the pretraining of the RNN is minimal. As a result, the initial generation of CNS drugs is rare, which negatively impacts the subsequent adversarial and reinforcement learning training. Furthermore, the discriminator is prone to being biased by the majority class, further reducing the likelihood of CNS drug generation.

To address this issue, we employed an oversampling technique during mixed training by tripling the sampling frequency of CNS drug data. This approach significantly improved the yield of CNS drugs, reaching over 5\%, which is substantially higher than training without oversampling (Figure 4A). However, further increasing the yield remains challenging due to two main reasons: (1) molecules of other classes also receive rewards, thus, the model tends to learn majority-class features to maximize local reward when sample distribution is imbalanced; (2) the limited number of CNS drug samples results in a less effective classifier during pretraining, as illustrated in Figure S4.

To further evaluate the quality of the generated compounds, we employed CNS\_MPO as a metric (Figure 4B). The average CNS\_MPO score of CNS drugs in the training set was 4.31. Molecules generated by ADSeqGAN without oversampling, as well as those produced by CLM and MolGEN, exhibited CNS\_MPO distributions comparable to that of the training set. However, we observed a decrease in CNS\_MPO scores when ADSeqGAN was trained with 3× oversampling. This decline may be attributed to the tendency of GANs to converge to a local optimum. Nonetheless, the decrease is acceptable, as ADSeqGAN with oversampling demonstrated a notably higher degree of molecular novelty compared to the other two few-shot models (Figure 4C). Furthermore, when the generated molecules were classified using our pre-trained random forest auxiliary discriminator, ADSeqGAN yielded significantly more molecules that matched the classifier’s criteria than the other models. It is important to note that the CNS\_MPO score and the random forest classifier serve different purposes: CNS\_MPO is designed to quantify the overall drug-likeness of CNS compounds, whereas the random forest model is trained to distinguish between CNS and non-CNS drugs. This difference in objectives may lead to a separation of shared features between CNS and non-CNS drugs that are otherwise embedded in the CNS\_MPO metric, thereby contributing to the observed reduction in CNS\_MPO scores.

To further validate the effectiveness of the generated molecules, we conducted molecular docking-based virtual screening against two CNS drug targets: 4MM5 \cite{Wang2013} and 6AWP \cite{Coleman2017}. These experiments were designed to evaluate the binding ability of the generated compounds. A virtual screening dataset comprising 6,400 samples was constructed using a generator trained under optimal performance conditions. After docking, we chose top hits to do case studies (Figure 4D, S6B and Table S1, S2). The 4MM5 structure represents the complex of the antidepressant sertraline with LeuBAT. Among the generated compounds, we identified molecules with structures highly similar to sertraline. Docking results showed that the fluorophenyl group in the generated sample aligned closely with the dichlorophenyl moiety in sertraline, occupying nearly the same position within the binding pocket. Moreover, the secondary amine group of the sample molecule spatially overlapped with the amine nitrogen atom of sertraline. Notably, the docking score of the generated sample was slightly better than that of sertraline, suggesting comparable or enhanced binding affinity. We further used SiteAF3\cite{Tang2025} to study binding poses and local interaction. In the 4MM5 system, ground-truth hydrogen bond distances were 4.2 Å and 3.4 Å, while our generated molecules exhibited 2.8 Å and 3.2 Å, showing superior binding ability. Similarly, 6AWP is the complex of the selective serotonin reuptake inhibitor (SSRI) fluvoxamine with the ts3 human serotonin transporter. Both the sample molecule and fluvoxamine featured a branched, tripodal structure. Docking analysis indicated that the amine group in the generated molecule occupied nearly the same spatial position as that in fluvoxamine and formed hydrogen bonds with the protein. In addition, the methoxy and aromatic side chains of both molecules exhibited similar orientations. Despite having comparable molecular sizes, the generated compound exhibited a higher docking score than fluvoxamine. Similar to 4MM5, the terminal amino groups of the molecules in 6AWP also capture similar hydrogen bond interactions. These findings support the capability of the ADSeqGAN model to effectively expand the chemical space of CNS-active molecules.

Overall, both ADSeqGAN itself and combined with oversampling provide a novel strategy for generating molecules from small-sample datasets, demonstrating its potential for addressing data scarcity in drug discovery.

\subsection{Expand ADSeqGAN Framework to CB1 ligand generation}

To illustrate the versatility of ADSeqGAN, we showed how well ADSeqGAN perform in generating novel molecular binders for a specific drug target, here it is CB1. We have collected about three thousand CB1R ligands which have measured inhibition constant ki values. We first constructed a classification model using the MACCS fingerprint as the descriptors. The molecular dataset is roughly balanced in terms of the numbers of the active and inactive compounds if we applied 1 µM as the threshold. For the generated molecules, we applied a set of drug likeness filters including the QED score and Lipinski’s Rule of 5 to filter out those non-druglike molecules. We have reported that the Glide docking score is a poor predictor for CB1R, thus, applied a target-specific scoring function, ligand-residue interaction profile-scoring function (LRIP-SF),\cite{Ji2021} to prioritize the designed molecules. As detailed in the Supporting Text, the LRIP-SF achieved an encouraging scoring power with the root-mean-square error of 1.27 kcal/mol and ranking power with correlation coefficient of 0.64.  We then applied the established LRIP-SF scoring function to predict the binding affinities of the designed druglike molecules. If we applied a threshold of -8.1854 kcal/mol, which corresponds to 1 µM, to determine if a compound is active or inactive, 32.8\% of designed molecules belong to the active group. This is an encouraging performance as the hit rate surpasses most CB1R-focused compound libraries not to mention the general-purpose screening libraries. As a conclusion, ADSeqGAN is able to generate novel and druglike molecules for a specific drug target.

\section{Conclusion}
In this study, we proposed ADSeqGAN, a novel sequence-based GAN framework incorporating auxiliary discriminators for small-sample molecular generation. By integrating a pretrained classifier as an additional discriminator, ADSeqGAN improves the generation of specific molecular classes, i.e., nucleic acid binders, CNS drugs and CB1 ligands. Our results demonstrate that ADSeqGAN outperforms traditional non-pretrained generative models in terms of molecular validity, diversity, and class specificity.

Through a combination of MLE pretraining generator, Wasserstein loss, and data augmentation techniques such as oversampling, ADSeqGAN effectively addresses mode collapse and enhances the generative process. The model exhibits strong label responsiveness, successfully differentiating nucleic acid binders from protein binders and generating CNS drugs at a significantly higher rate than standard approaches. Furthermore, docking experiments confirm the ability of ADSeqGAN in generating high-quality molecules, thus it has a great application in enlarging compound library for nucleic acid and CNS drug discovery.

Future work will focus on incorporating molecular scaffold information and SMILES syntax rules into the generation process to improve the success rate of valid molecule generation. We also aim to refine training strategies to further optimize molecular properties and integrate more advanced reinforcement learning techniques to enhance chemical space exploration.

\section{Supporting Information}

Additional experimental details on nucleic acid binders, central nervous system drugs, and CB1R ligands; evaluation of random forest classifiers built on physical chemical descriptors, MACCS and CDDD; PCA analysis; docking and binding pose comparisons; scaffold and functional group analysis; case study on CB1R with LRIP-SF; computational details of ADSeqGAN training and generation. (PDF)

\section{Acknowledgment}
This work was supported by funds from the National Institutes of Health (R01GM147673, R01GM149705) and the National Science Foundation (1955260). The authors would like to thank the computing resources provided by the Center for Research Computing (facility RRID: SCR\_022735) at the University of Pittsburgh (NSF award number OAC-2117681), and the Pittsburgh Supercomputer Center (grant number BIO210185)

\section{Data and Software Availability}
All the training datasets, code and parameters are available online at GitHub:  \noindent\texttt{https://github\allowbreak.com/ClickFF/ADSeqGAN} and
\texttt{https://github.com/HaCTang/ADSeqGAN}.

\section{AUTHOR INFORMATION }

\subsection{Corresponding Author} 

Junmei Wang - Department of Pharmaceutical Sciences and Computational Chemical Genomics Screening Center, School of Pharmacy, University of Pittsburgh, Pittsburgh, PA 15261, USA. Email: juw79@pitt.edu 

\subsection{Other Authors} 

Haocheng Tang - Department of Pharmaceutical Sciences and Computational Chemical Genomics Screening Center, School of Pharmacy, University of Pittsburgh, Pittsburgh, PA 15261, USA. Email: hat170@pitt.edu 

Jing Long - School of Software \& Microelectronics, Peking University, Beijing, 100871, CN. Email: jing.long0926@gmail.com 

Beihong Ji - Department of Pharmaceutical Sciences and Computational Chemical Genomics Screening Center, School of Pharmacy, University of Pittsburgh, Pittsburgh, PA 15261, USA. Email: bej22@pitt.edu

\subsection{Author Contributions} 
J.W. and H.T. designed the research. H.T., L.J. and B.J. performed the research. H.T. analyzed the data and wrote the manuscript. J.W. provided computing resources. All authors reviewed the manuscript.

\subsection{Conflict of Interest}
The authors declare no competing financial interests.

\let\thefootnote\relax\footnotetext{*Corresponding author: Junmei Wang (\texttt{JUW79@pitt.edu})}
    
\bibliography{references}

\end{document}